\documentclass[a4paper, 10pt, conference]{ieeeconf}      

\IEEEoverridecommandlockouts                              

\usepackage{amsmath}
\usepackage{siunitx}	
\DeclareMathOperator*{\argmin}{arg\,min}

\newtheorem{rem}{Remark}

\usepackage{array}		
\newcolumntype{P}[1]{>{\centering\arraybackslash}p{#1}}

\usepackage{graphicx} 
\usepackage[caption=false, font=footnotesize]{subfig}

\usepackage{tikz}		
\usepackage{pgfplots}
\pgfplotsset{compat=newest,compat/show suggested version=false}

\usepgfplotslibrary{fillbetween}

\usepackage{xcolor}

\definecolor{aoeng}{rgb}{0.0, 0.5, 0.0}               
\definecolor{arylideyellow}{rgb}{0.91, 0.84, 0.42}    

\title{\LARGE \bf
On Maximizing Lateral Clearance of an Autonomous Vehicle in Urban Environments
}

\author{
Francesco Seccamonte, Juraj Kabzan and Emilio Frazzoli
\thanks{The authors are with nuTonomy, an Aptiv company.\newline {\tt\small \{francesco, juraj, emilio\}@nutonomy.com.} }%
}

\begin{document}

\maketitle
\thispagestyle{empty}
\pagestyle{empty}

\begin{abstract}

We consider the problem of maximizing distance to road agents for a self-driving car.
To this extent, we employ a Model Predictive Control (MPC) approach for the steering tracking control of an Autonomous Vehicle (AV).
Specifically, we first present a traditional MPC controller, which is then extended to encode the clearance maximization goal by manipulating its cost function and constraints.
We provide insights on the additional information needed to achieve such goal, and how this modifies the structure of the original controller.
Furthermore, a connection between commonly used safety metrics and clearance to road users is established.
We implement the MPC controller using two off-the-shelf numerical solvers, assessing its computational feasibility.
Finally, we show experimental results of the proposed approach on public roads in Boston and in Singapore.

\end{abstract}

\section{INTRODUCTION}

The execution of actions in a robotics system is often decomposed into a Planning and Decision-making layer (in the following, simply the planner), and a Controller module.
The former is responsible for choosing a maneuver to be performed and for imposing restrictions in speed, time, and space on how this maneuver shall be executed.
The latter is responsible for the optimal execution of the maneuver within the bounds supplied by the planner, possibly taking into account the physical limitations of the dynamical system as well.
Such decomposition is common in Autonomous Vehicles (AVs), where the actions taken must obey specified rules and also represent a safe and socially acceptable behavior.
The decision-making process is obviously based on the output of perception and prediction modules, which provide information about the current and predicted state of the world.
Sampling-based motion planning algorithms \cite{lavalle_2006} have become extremely popular in the AV domain as a solution for the decision-making problem, in particular the ones able to handle non-holonomic constraints, such as RRT \cite{Lavalle98rrt} and its asymptotically optimal variant RRT* \cite{frazzoli-rrt}.
In an urban environment, such algorithms take on great importance, as they allow to encode safety rules, e.g., in a minimum-violation fashion \cite{sample-based-planning1, sample-based-planning2}.
On the downside, navigating dense urban environments requires a thorough refinement of the trajectory to be executed.

\begin{figure}[ht]
  \begin{center}
    \includegraphics[width=\linewidth]{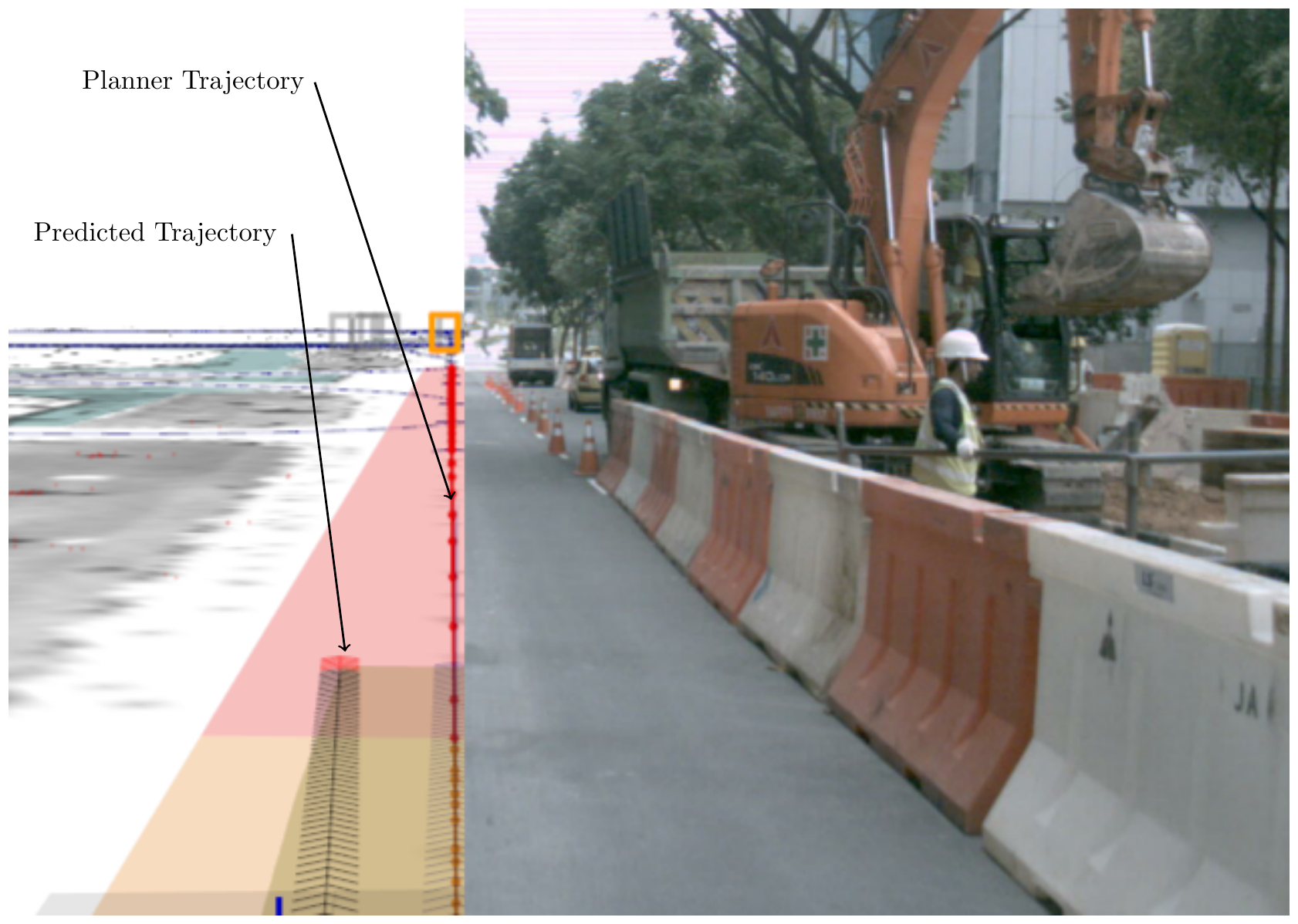}
    \caption{The nuTonomy R\&D platform navigating a road in Singapore. The AV ``sees'' a construction zone (right), and deviates from the reference to keep a higher distance (black arrows on the left).}
    \label{img:biasing-construction}
    \end{center}\vspace{-0.5cm}
\end{figure}

Fig. \ref{img:biasing-construction} shows a narrow road traveled by our AV, with a construction zone on the right side.
In such scenario it is extremely helpful to finely pick a path farther from the construction area.
Indeed, being too close too it and interacting with pedestrians potentially moving outside of the area unexpectedly, or even construction material falling off might endanger safety of the AV's passengers and of other road users.

Refining the driven path at the sampling-based planning level results in a large number of samples needed, dramatically increasing its computational complexity.
While speeding up motion planning algorithms is an active area of research \cite{valerio-pruning}, another possibility is lowering their complexity by deferring to lower layers (such as the controller) the refinement of trajectories for achieving additional goals.
The second approach is the one we adopt for maximizing lateral clearance, as optimization over a continuous domain can be easily attained via gradient-based methods, which are widely used in control applications.

In a nutshell, such architecture\footnote{Please note that the functionality described is not necessarily representative of current and future products by nuTonomy, Aptiv and their partners.
The scenarios presented are simplified for the purposes of exposition.
The examples discussed are illustrative of the philosophy but not the precise specification we use.
Moreover, we gloss over the extensive verification and validation processes that are performed before driving on public roads.}
considers the combinatorial aspect of the decision-making process (e.g., to overtake a car or to stop for it) tackled at the higher decision-making level, while the local refinement of the trajectory is handled in the controller module.
The goal of this paper is to present how the latter is performed; details on the planner employed, or on the perception and prediction modules used, are outside the scope of this paper.

\subsection{Related work}

Model Predictive Control (MPC) is nowadays the standard for model-based optimal control.
It can be used for both regulation and tracking problems, and it allows to explicitly account for constraints on the system.
Applications in the Autonomous Driving domain range from motion planning for miniature race cars \cite{Liniger-cars}, to the investigation of different models suitable for an AV driving task on full size vehicles \cite{borrelli-mpc-models}, and more recently an MPC-based planner which includes also a collision avoidance system \cite{bmw-mpc}.
This last system, experimentally demonstrated also on full sized vehicles, does not lead the AV to choose a safer path (that is, a path with larger clearance to obstacles): Avoiding the obstacle by $\SI{0.1}{\metre}$ or $ \SI{10}{\metre}$ is considered equivalent.

Inducing the choice of a safer path can be achieved by modifying a traditional MPC architecture to specify additional objectives; Such approach is a fairly recent trend for autonomous systems.
The authors in \cite{Nageli:IEEE,Nageli:2017siggraph} employ this framework to maximize targets visibility in drone cinematography; authors in \cite{PAMPC-Scaramuzza} similarly encode the additional goal of keeping a point of interest in sight for a UAV.
The MPC frameworks therein implemented serves as both planner and controller, given that the drone can simply move from point $A$ to point $B$ without any specific rule to obey.

In \cite{hans-ITSC}, the authors extend an MPC-based planner to account for visibility maximization while overtaking a static obstacle in an urban environment.
Rule compliance is taken into account by means of a state machine which selects the maneuver that the MPC shall deliver.
The paper includes simulation results showing the effectiveness of the approach in a scenario including only a single parked car.
Extensions to more sophisticated scenarios can become very challenging; also, it is unclear how many agents such approach can handle simultaneously.

\subsection{Contribution}
Compared to the aforementioned approaches, our work differs in the way responsibilities are assigned to our AV's modules:
The planning layer needs to ensure rule compliance, for example according to \cite{RulebooksICRA}, and provide the controller a reference trajectory to be followed, as well as a rule compliant region in the phase space (in the following, the ``tube'').
Such region empowers the controller with some freedom, as it is both collision-free and rule compliant.
The proposed controller is able to track the reference trajectory, and at the same time to retain the capability of changing the exact driven path, that is, \textit{how} the AV performs the maneuver.
It is worth mentioning that the proposed scheme is compatible with any planning algorithm, as no assumptions are made on its architecture (e.g., state machine or graph based).
At the controller level, comfort and tracking objectives are then considered jointly with safety related goals, that is, maximizing clearance to other agents.
This combination of goals requires a trade-off between tracking objective and clearance maximization goal, resulting in what we refer to as \textit{biasing} \cite{nut-lane-biasing}.

\begin{figure}[thpb]
	\begin{center}
    \parbox{3in}{\begin{tikzpicture}

    \coordinate (traj_start) at (1.0, 0.55);    
    \coordinate (traj_end) at (7.5, 0.55);      
    
    \draw[thick](0,2.2) -- (7.5,2.2);         
    \draw[dashed](0, 1.1) -- (7.5, 1.1);
    \draw[thick](0,0) -- (7.5,0);
    
    \draw (0.2, 0.35) rectangle (1.0, 0.75);
    \draw[fill=lightgray] (3.6, 0.075) rectangle (4.4, 0.475);

    \draw (4, 0.275) node {\small \( O \)};
    \draw (0.6, 0.55) node {\small \( V \)};

    \draw [blue] plot [smooth] coordinates
    { (traj_start) (1.5, 0.62) (4.0, 1.2) (7.0, 0.62) (traj_end) }; 
    \draw [red, ->] plot [smooth] coordinates
    { (traj_start) (1.5, 0.65) (3.3, 1.4) (4.0, 1.53) (4.7, 1.4) (7.0, 0.65) (traj_end) };  

    \draw [aoeng, name path = tube_up] plot [smooth] coordinates
    { (1.0, 0.95) (1.5, 1.02) (4.0, 1.6) (7.0, 1.02) (7.5, 0.95) }; 
    \draw [aoeng, name path = tube_low] plot [smooth] coordinates
    { (1.0, 0.25) (1.5, 0.3) (4.0, 0.75) (7.0, 0.3) (7.5, 0.25) };  

    \tikzfillbetween[of=tube_up and tube_low, on layer=]{aoeng, opacity=0.15};

\end{tikzpicture}
 }
  	\caption{Qualitative paths during an avoidance maneuver (left-hand traffic): planner's reference (blue) and controller's biased (red). The ``tube'' consists of the shaded green region.}
    \label{img:traj}
  	\end{center}
\end{figure}
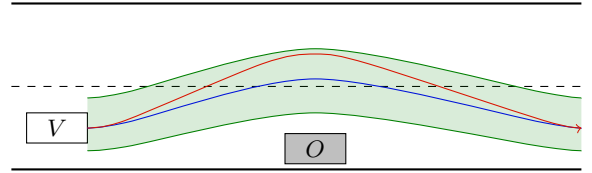

Thanks to this additional functionality, the controller does not only follow the planner's trajectory, but is also directly aware of perceived objects.
An illustration of this behavior is depicted in Fig. \ref{img:traj}.

Similarly to related approaches in the literature, an MPC control framework is chosen mostly for its ability to handle constraints, which are a key component to enable biasing functionality.

The contribution of this paper is threefold: Firstly, the MPC control framework is presented (Section \ref{sec:formulation}), together with its clearance maximization extension (Section \ref{sec:biasing}).
Insights on how to allow the MPC layer to consume objects information, as well as a connection between clearance to such objects and commonly used safety metrics, are given in Section \ref{sec:constraints}.
Secondly, the resulting optimization problem is implemented using two different off-the-shelf numerical solvers, and its computational tractability is assessed (Section \ref{sec:res}).
Finally, extensive experimental results are shown using the nuTonomy R\&D platform, demonstrating the validity of the proposed approach in real-world scenarios.

\section{MATHEMATICAL NOTATION}
\label{sec:math_prel}
We denote vectors with bold letters, and the column vector given by $[ \mathbf{x}^T, \mathbf{y}^T ]^T$ as $ (\mathbf{x}, \mathbf{y}) $.
The time derivative of the function $f(t)$ is denoted as $\dot{f}(t)$.
The sampling time of the system is denoted as $T_s$.
We use $x_k=x(kT_s)$ to represent the $k$-th prediction step of variable $x$, where $k = \{ 0, 1, \ldots, N\}$.

\section{MPC FORMULATION}
\label{sec:formulation}

In this section the design of the lateral tracking controller is presented.

\subsection{Vehicle's model}

The first component of an MPC algorithm is a model of the dynamical system to be controlled.
Since the operational domain considers an AV driving in urban scenarios, the maximum achievable speed is low (around $\SI{14}{\metre\per\second}$ depending on the regulations); similarly, acceleration is low as well.
As highlighted in \cite{borrelli-mpc-models}, in low speed and acceleration regimes the usage of a higher fidelity model does not provide any substantial benefit over a kinematic one.
Hence, we choose to model the AV with a kinematic bicycle model \cite{schramm-vehicle-dynamics} \cite{Laumond-book}.
Denoting by $x,y, \theta$ the position and heading of the AV at the center of the rear axle in a global inertial frame, by $v$ the speed at the rear axle, by $\delta$ its ground steering angle, we express the heading angle dynamics as a function of speed and curvature at the rear axle $\kappa$, using the relation $\kappa(t) = \tan(\delta(t))/b$, with $b$ denoting the wheelbase.
Given the physical limits of the system, we can directly recover the ground steering angle $\delta(t)$ as $\delta(t) = \arctan(\kappa(t) b)$.
Moreover, we model the actuation dynamics as a first-order dynamical system with time constant $\tau$ (with $\tau$ being a parameter identified using standard system identification techniques \cite{Ljung-sys-id}), hence adding the additional variable $\kappa_{\text{des}}$ denoting the desired curvature.
This results in the following system of differential equations:
\begin{subequations}
	\label{sys_dyn_ct}
	\begin{align}
	& \dot{x}(t) 				= v(t) \cos(\theta(t)) \, ,\label{x_dyn}\\
	& \dot{y}(t) 				= v(t) \sin(\theta(t)) \,	,\label{y_dyn}\\
	& \dot{\theta}(t) 			= v(t) \kappa(t) \, , 	 \label{theta_dyn}\\
	& \dot{\kappa}(t) 			= \frac{1}{\tau}(\kappa_{\text{des}}(t) - \kappa(t)) \, ,\\
	& \dot{\kappa}_{\text{des}}(t)  	= u(t) 	\, .		  \label{kappa_dyn}
	\end{align}
\end{subequations}

\begin{figure}[thpb]
  \begin{center}
    \def\svgwidth{0.7\columnwidth}
    \framebox{\parbox{3in}{\centering 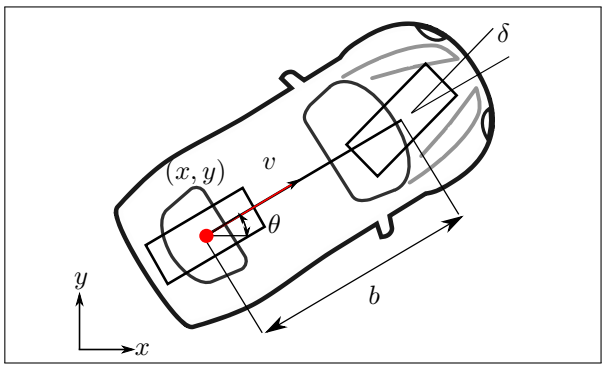}}
    \caption{Kinematic bicycle model of a car.}
    \label{img:car}
    \end{center}
\end{figure}

In order to be employed in an MPC scheme, the continuous time differential equations \eqref{sys_dyn_ct} need to be discretized, and the Explicit Runge-Kutta numerical scheme is employed.
Stacking together the state variables, we write the state vector $\mathbf{x} = \left( x, y, \theta, \kappa, \kappa_{\text{des}}\right)$. Eq. \eqref{kappa_dyn} shows the input $u$ being the desired curvature rate.
The model is depicted in Fig. \ref{img:car}.

\begin{rem}
The proposed scheme is a steering tracking controller, hence the speed $v(t)$ in Eq. \ref{sys_dyn_ct} is an uncontrollable parameter and not an optimization variable.
\end{rem}

\subsection{Cost function}

The proposed control architecture aims at addressing a tracking problem, hence the objective function is written in terms of tracking error.
By denoting the reference trajectory vector at stage $k$ as $\mathbf{z}_k = \left( \bar{x}_k, \bar{y}_k, \bar{\theta}_k, \bar{\kappa}_k\right)$, the cost to be minimized is the sum of the tracking error at each prediction step $k = 0, \dots, N$, with $N$ being the prediction horizon length (in time steps), that is
\begin{equation}
\label{mpccost_formal}
J_{\text{nom}} = \sum_{k=0}^{N-1} \left( q(\mathbf{x}_k, \mathbf{z}_k ) + r(u_k)\right) + p(\mathbf{x}_N, \mathbf{z}_N ) \, .
\end{equation}

The quantity $\mathbf{z}_k, k = \{ 0, \ldots, N\}$ is obtained by discretizing the continuous planner's reference trajectory, using the predicted speed $v_k$ the AV will drive at the corresponding time instants (which comes from a different module).

Similarly to what is performed in contouring control \cite{Liniger-cars}, we compute the lateral error at each prediction step $k$ as 
\begin{equation*}
e_{\text{lat},k} = -(x_k - \bar{x}_k)\sin(\bar{\theta}_k) + (y_k - \bar{y}_k)\cos(\bar{\theta}_k) \, ,
\end{equation*}
and penalize it in the cost function.

Hence, by employing a quadratic cost function, the term $q(\mathbf{x}_k, \mathbf{z}_k )$ corresponds to
\begin{equation*}
q_1 e_{\text{lat},k}^2 + q_2(\theta_k - \bar{\theta}_k)^2 + q_3(\kappa - \bar{\kappa}_k)^2 \, ,
\end{equation*}
with $q_1, q_2, q_3 \geq 0$.
The term $p(\mathbf{x}_N, \mathbf{z}_N )$ is structured analogously (with different weights only), and the stage input cost is $r(u_k) = R u_k^2$, $R > 0$.

\subsection{Constraints}

The following constraints are imposed on the system:
\begin{subequations}
	\label{constraints}
	\begin{align}
	& \underline{u} \leq u_k \leq \overline{u}                   		&\forall \, &k = \{ 0, \ldots, N-1\} 	\, ,\label{constr:input}\\
	& \underline{\kappa} \leq \kappa_k \leq \overline{\kappa}    		&\forall \, &k = \{ 0, \ldots, N\} 		\, ,\label{constr:kappa}\\
	& \underline{e_{\text{lat},k}} \leq e_{\text{lat},k} \leq \overline{e_{\text{lat},k}}    &\forall \, &k = \{ 0, \ldots, N\} \, .	\label{constr:tube}
	\end{align}
\end{subequations}

Constraints \eqref{constr:input} and \eqref{constr:kappa} bound curvature rate and curvature taking into account physical limits; \eqref{constr:tube} constrains the (signed) lateral error to lie within time-varying bounds. In order to ensure feasibility, the inequalities \eqref{constr:tube} are expressed as soft constraints, through the addition of a slack variable $\varepsilon_k , \varepsilon_k \geq 0$, which is also penalized in the cost function by means of linear and quadratic penalties \cite{mpc-book}.

\section{Maximizing lateral clearance within MPC}
\label{sec:biasing}

In this section, we extend the previously derived MPC controller by manipulating its cost function and constraints.

\subsection{Safety related auxiliary variable}

Since we aim at maximizing lateral clearance in appropriate circumstances (e.g., in the proximity of agents whose motion might change unpredictably), as well as minimizing the tracking error, a natural approach is to extend the cost function by adding suitable terms.
The simple and intuitive observation is the following: The higher distance a vehicle has with respect to other objects, the lower the likelihood of incurring accidents, hence the higher safety.
More formally speaking, incrementing clearance increases the commonly used Time-To-React (TTR) metric quantity \cite{time-to-c}.

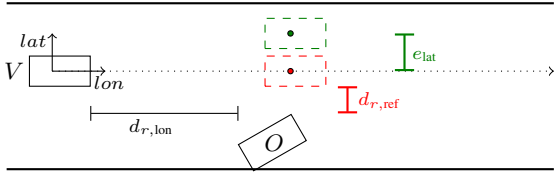
\begin{figure}[thpb]
	\begin{center}
    \parbox{3in}{\begin{tikzpicture}
    
    \draw[thick](-0.2,2.2) -- (7.0,2.2);                
    \draw[dotted, ->](0.4, 1.3) -- (7.0, 1.3);          
    \draw[thick](-0.2,0) -- (7.0,0);                    

    \draw [] (0.1, 1.1) rectangle (0.9, 1.5);           
    \draw (-0.1, 1.3) node {\small \( V \)};            
    \draw [shift={(3.5 cm, -0.8 cm)},rotate=30] (0.0, 0.9) rectangle (0.8, 1.3);   
    \draw (3.32, 0.35) node {\small \( O \)};           

    \draw [->] (0.4, 1.3) -- (1.1, 1.3);                
    \draw (1.15, 1.15) node {\scriptsize \( lon \)};    
    \draw [->] (0.4, 1.3) -- (0.4, 1.8);                
    \draw (0.15, 1.7) node {\scriptsize \( lat \)};     

    \draw [dashed, red, shift={(3.1 cm, 0.0 cm)}] (0.1, 1.1) rectangle (0.9, 1.5);          
    \draw [dashed, aoeng, shift={(3.1 cm, 0.2 cm)}] (0.1, 1.4) rectangle (0.9, 1.8);        
    \filldraw[fill=red] (3.535, 1.3) circle (1pt);                                          
    \filldraw[fill=aoeng] (3.535, 1.8) circle (1pt);                                        
    
    \draw [thick, red, shift={(0.75 cm, 0.0 cm)}, |-|] (3.535, 0.74) -- (3.535, 1.1);       
    \draw [red] (4.7, 0.9) node {\scriptsize \( d_{r,\text{ref}} \)};                              
    \draw [thick, aoeng, shift={(1.5 cm, 0.2 cm)}, |-|] (3.535, 1.1) -- (3.535, 1.6);       
    \draw [aoeng] (5.3, 1.5) node {\scriptsize \( e_{\text{lat}} \)};                              

    \draw [|-|] (0.9, 0.74) -- (2.85, 0.74);                    
    \draw (1.7, 0.55) node {\scriptsize \( d_{r,\text{lon}} \)};         

\end{tikzpicture}}
  	\caption{Both longitudinal $d_{r,\text{lon}}$ and lateral $d_{r,\text{ref}}$ distances are computed from the discretized planner's reference (axes origin) to the object $O$ and account for the car's and object's footprints. Biasing (dashed green rectangle) increases the actual lateral distance by $e_{\text{lat}}$.}
    \label{img:clearance}
  	\end{center}
\end{figure}

To this extent, we introduce the auxiliary optimization variable $s$ representing achievable safety, that is, an empirical measure of the attainable safety to be correlated to the clearance given to objects.
For ease of exposition, the case with one object on the right of the reference trajectory is considered first (illustration in Fig. \ref{img:clearance}); the case of one object on the left is analogous, and how to handle multiple road agents is described in Section \ref{subsec:multiple}.

We denote by $d_{r,\text{ref}}$ the lateral distance from the reference to the object on its right.
Such distance accounts for the car's footprint, and is non-negative.
The actual distance from the AV to the agent is denoted as $d_r$, and is computed as $d_r = d_{r,\text{ref}} + e_{\text{lat}}$.
The lateral error $e_{\text{lat}}$ is a signed quantity.
For this reason, the case of the object being on the left side results in $d_l = d_{l,\text{ref}} - e_{\text{lat}}$.
Fig. \ref{img:clearance} depicts such quantities.

The cost function \eqref{mpccost_formal} is extended via
\begin{equation}
	\label{mpccost_biasing}
	J_{\text{bias}} = \sum_{k=0}^{N} \left( s_k - s_{\text{target}} \right)^2 \; ,
\end{equation}
with $s_{\text{target}}$ being a constant representing the target safety.

Safety and lateral distances are linked through the clearance-dependent safety constraint function $f_s(d)$.
Hence, the following constraint is added:
\begin{equation}
	\label{constraint:f_s}
	s \leq f_s(d_r) + s_{r,\text{lon}}\, .
\end{equation}

The safety term $s_{r,\text{lon}}$ is related to the longitudinal distance $d_{r,\text{lon}}$ via the function $f_{\text{lon}}(d_{\text{lon}})$.
Such function is non-increasing in $(-\infty,0)$ and non-decreasing in $[0, \infty)$, with $f_{\text{lon}}(0) = 0$.
Since $d_{r,\text{lon}}$ is an uncontrollable real time parameter, the term $s_{r,\text{lon}}$ is computed outside of the MPC controller.

The function $f_s(d)$ is a function of the lateral clearance $d$ and also non-decreasing; additional details are given below.

\subsection{Shaping the clearance-dependent safety function}
\label{sec:f_s-function}

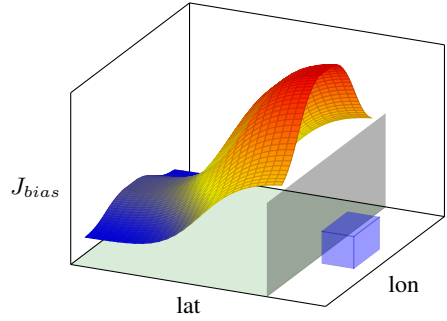
\begin{figure}[thpb]
  \begin{center}
      \parbox{3in}{\begin{tikzpicture}

	\pgfmathsetmacro{\Xmin}{-4}
	\pgfmathsetmacro{\Xmax}{1.2}
	\pgfmathsetmacro{\Ymin}{-8}
	\pgfmathsetmacro{\Ymax}{8}
	\pgfmathsetmacro{\Zmin}{-0.2}
	\pgfmathsetmacro{\Zmax}{2}
	
	\begin{axis}[
        xtick=\empty,
        ytick=\empty,
        ztick=\empty,
		xlabel=lat, ylabel=lon, zlabel=$J_{bias}$,
		zlabel style={rotate=-90},	
		xmin=\Xmin,
		xmax=\Xmax,
		ymin=\Ymin,
		ymax=\Ymax,
		zmin=\Zmin,
		zmax=\Zmax,
		small
		]


	\pgfmathsetmacro{\Xstart}{\Xmin}
	\pgfmathsetmacro{\Xend}{0}

	\pgfmathsetmacro{\targetX}{1}
	\pgfmathsetmacro{\gainX}{2}
	\pgfmathsetmacro{\offsetX}{-2}

	\pgfmathsetmacro{\targetY}{0.5}
	\pgfmathsetmacro{\gainY}{-1}
	\pgfmathsetmacro{\offsetY}{4}

	\pgfmathsetmacro{\Xline}{0}	

	\pgfmathsetmacro{\valueAtObst}{\targetY/(1 + exp(\gainY*(1 - \offsetY))) - \targetY}
	
	\addplot3[name path=uppath, opacity=0.1, fill opacity=0.4, domain y=\Ymin:\Ymax] (\Xline,y,\Zmin);
	\addplot3[name path=botpath, opacity=0.1, fill opacity=0.4, domain y=\Ymin:\Ymax] (\Xstart,y,\Zmin);
	\addplot [aoeng, opacity=0.15] fill between[of=uppath and botpath];


	\addplot3 [
		surf,
		no markers,
		domain=\Xstart:\Xend,
		domain y=1:6,
		] { (\targetX/(1 + exp(\gainX*(x - \offsetX))) - \targetX + \targetY/(1 + exp(\gainY*(y - \offsetY))) - \targetY)^2 };

	\addplot3 [
		surf,
		no markers,
		domain=\Xstart:\Xend,
		domain y=-6:-1,
		] { (\targetX/(1 + exp(\gainX*(x - \offsetX))) - \targetX + \targetY/(1 + exp(\gainY*(-y - \offsetY))) - \targetY)^2 };

	\addplot3 [
		surf,
		no markers,
		domain=\Xstart:\Xend,
		domain y=-1:1,
		] { (\targetX/(1 + exp(\gainX*(x - \offsetX))) - \targetX + \valueAtObst)^2 };

	\addplot3[name path=uppath1, opacity=0.1, fill opacity=0.4, domain y=\Ymin:\Ymax] (\Xline,y,\Zmin);
	\addplot3[name path=upline, opacity=0.0, fill opacity=0.4, domain y=\Ymin:\Ymax] (\Xline,y,1);
	\addplot [
    opacity=0.3,
    shader=interp,
    ] fill between[of=upline and uppath1];

	\addplot3 [
		rotate around z=0,
		only marks,
		scatter,
		mark=cube*,
		mark size=12,	
		opacity=0.4,
		] coordinates {(0.5, 0, 0)};

	\end{axis}

\end{tikzpicture}}
    \caption{Value of the cost function $J_{\text{bias}}$ for the example in Fig. \ref{img:clearance}. The green shaded area defines the tube, while the blue cube represents the obstacle.}
    \label{plot:f_s}
    \end{center}
\end{figure}

Depending on how the clearance-dependent safety function is shaped, the extent to which the car deviates from the reference will be different. We can use functions taking values in the range $\left[ 0, s_{\text{target}} \right]$: by relating safety to Time-To-React (TTR), a negative TTR (hence safety) relates to unavoidable collision states only.
However, thanks to the Planner outputting collision-free tubes only, we can neglect the cases where TTR $ < 0$. Moreover, safety cannot grow unbounded, since after a certain threshold a higher TTR does not provide any additional benefit. We employ sigmoid-like functions, e.g., $f_s(d) = \frac{s_{\text{target}}}{1 + e^{-a(d-b)}}$, for both the lateral and longitudinal terms. Figure \ref{plot:f_s} depicts the value of the cost function \eqref{mpccost_biasing} (for one time step only) associated to the scenario drawn in Fig. \ref{img:clearance}.

\subsection{Control law}

Putting together the dynamics equations \eqref{sys_dyn_ct} (in the form of equality constraints), the cost functions \eqref{mpccost_formal} and \eqref{mpccost_biasing} and the constraints \eqref{constraints} and \eqref{constraint:f_s} (whose resulting feasible sets are compactly denoted as $\tilde{\mathcal{X}}$ and $\mathcal{U}$), we extend the state variables by adding the safety term, that is, $\tilde{\mathbf{x}} = \left( \mathbf{x}, s \right)$. Note that $s$ has no dynamics, hence the equality constraints deriving from \eqref{sys_dyn_ct} are unchanged. Finally, denoting the sequence of inputs $\{u_0, \ldots , u_{N - 1}\}$ at time steps $\{0,\ldots, N-1\}$ as $\mathbf{u}$ we can write the full optimization problem:
\begin{subequations}
  \label{MPC_formulation}
  \begin{align}
    \underset{\mathbf{u}}{\text{min}} \quad & J_{\text{nom}} + \alpha J_{\text{bias}} \label{mpc_cost_full}\\
    \text{s.t.} \quad & \mathbf{x}_0 = \mathbf{x}(t) \, ,\label{prob:init}\\
    & \mathbf{x}_{k+1} = f(\mathbf{x}_{k}, u_{k}), 				&\forall \, &k = \{ 0, \ldots, N-1\} \, , 	\label{model_control}\\
    & \tilde{\mathbf{x}}_{k} \in \, \tilde{\mathcal{X}}_{k}, 	&\forall \, &k = \{ 0, \ldots, N\} \, , 	\label{state_constraints_compact}\\
    & u_{k} \in \, \mathcal{U}, 								&\forall \, &k = \{ 0, \ldots, N-1\} \; .	\label{input_constraints_compact}
\end{align}
\end{subequations}

The factor $\alpha \geq 0$ is a tuning parameter used to trade off between tracking performance and clearance maximization.
At each time step, problem \eqref{MPC_formulation} is initialized with the measured state $\mathbf{x}(t)$ as initial condition (constraint \eqref{prob:init}).
The new reference trajectory $\mathbf{z}_k$ as well as the distances $d_{l,\text{ref},k}, d_{r,\text{ref},k}, d_{l,\text{lon},k}, d_{r,\text{lon},k}, \forall k = \{0,\ldots, N\}$, are also updated.
An optimal solution of \eqref{MPC_formulation} is $J^{\star}$, with associated optimizer $\mathbf{u}^\star = \{u_0^\star, \ldots , u^\star_{N - 1}\}$.
Problem \eqref{MPC_formulation} is solved in a receding horizon fashion \cite{mpc-book}, and only the first input $u_0^\star$ is applied to the system.

\begin{rem}
Given that constraints \eqref{constr:tube} are soft, problem \eqref{MPC_formulation} is always feasible.
This comes from the fact that constraints \eqref{constr:kappa} are related to physical limits of the vehicle (hence always feasible provided a feasible initial condition \eqref{prob:init}), and constraints \eqref{constr:input} and \eqref{constraint:f_s} limit optimization variables with no dynamics (thus they can vary freely and instantaneously).
Such property is extremely important for the real-world deployment of the MPC controller.
\end{rem}

\section{CONSTRAINTS COMPUTATION}
\label{sec:constraints}

In this section we present details on the lateral error bounds (i.e., the ``tube'') and the bias-inducing terms.
The tracked objects information (measured and predicted for the future $k$ steps) are obtained from separate perception and prediction layers, which do not belong to the scope of this paper.

\subsection{Examples of tube computation}

The role of constraints \eqref{constr:tube} is of paramount importance: They are needed to specify the maximum lateral error the AV can tolerate due to biasing.
For this reason, such ``tube'' has to represent a collision-free region around the reference trajectory, where the car can freely move without endangering safety or violating specifications.
To this extent, all the available informations are used for tube derivation: offline (i.e., map-based) and online (i.e., perception-based).
Moreover, in some situations a narrower tube is advisable: this is the case, for instance, of the AV navigating an intersection, and is due to the fact that the lane boundaries are often not well defined, and other cars do not follow their prescribed paths but tend to ``cut''.
In such situations the tube is shrunk progressively as the AV approaches the maneuver requiring superior tracking performance to avoid abrupt swerves that would greatly impact comfort, or even endanger safety.

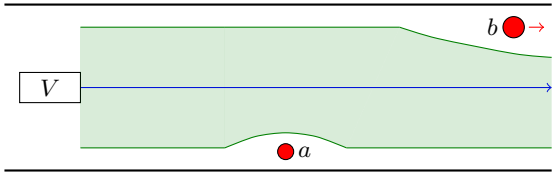
\begin{figure}[thpb]
	\begin{center}
    \parbox{3in}{\begin{tikzpicture}
    
    \draw[thick](-0.2,2.2) -- (7.0,2.2);            
    \draw[blue, ->](0.8, 1.1) -- (7.0, 1.1);        
    \draw[thick](-0.2,0) -- (7.0,0);                

    \draw (0.0, 0.9) rectangle (0.8, 1.3);          
    \draw (0.4, 1.1) node {\small \( V \)};         
    \filldraw[fill=red] (6.5, 1.9) circle (4pt);    
    \draw[red, ->](6.7,1.9) -- (6.9,1.9);           
    \filldraw[fill=red] (3.5, 0.25) circle (3pt);   

    \draw (6.22, 1.9) node {\small \( b \)};        
    \draw (3.75, 0.25) node {\small \( a \)};       

    \draw[aoeng, name path=tube_up1](0.8, 1.9) -- (2.7, 1.9);        
    \draw[aoeng, name path=tube_up_mid](2.7, 1.9) -- (5.0, 1.9);     
    \draw[aoeng, name path=tube_low1](0.8, 0.3) -- (2.7, 0.3);       
    \draw[aoeng, name path=tube_low2](4.3, 0.3) -- (7.0, 0.3);       

    \draw [aoeng, name path=tube_up2] plot [smooth] coordinates
    { (5.0, 1.9) (5.5, 1.75) (6.5, 1.55) (7.0, 1.5) };               

    \draw [aoeng, name path=tube_low_arc] plot [smooth] coordinates
    { (2.7, 0.3) (3.1, 0.45) (3.5, 0.5) (3.9, 0.45) (4.3, 0.3) };    

    \tikzfillbetween[of=tube_up1 and tube_low1, on layer=]{aoeng, opacity=0.15};        
    \tikzfillbetween[of=tube_up_mid and tube_low_arc, on layer=]{aoeng, opacity=0.15};  
    \tikzfillbetween[of=tube_up2 and tube_low2, on layer=]{aoeng, opacity=0.15};        

\end{tikzpicture}
 }
  	\caption{Planner's trajectory (blue) and collision-free tube (green) accounting for a static (a) and a moving (b) obstacles, as well as road boundaries.}
    \label{img:tube}
  	\end{center}\vspace{-0.0cm}
\end{figure}

An example of the tube is depicted as the green shaded area in Fig. \ref{img:tube}.

\subsection{Handling multiple objects}
\label{subsec:multiple}

In urban driving, it is not uncommon to have multiple road agents that the AV needs to account for.
To this extent we introduce the notion of \textit{most constraining object} on the left and on the right, $o_l$ and $o_r$, respectively: that is, at each prediction time step $k$ the object that constrains the safety variable $s_k$ the most, for a fixed $e_{\text{lat}} = 0$.
Note that the most constraining object at time step $(k+1)$ is not necessarily the same as at time step $k$, given its predicted motion.

Such objects are found via
\begin{equation}
	\label{eq:dist_objects}
	\begin{aligned}
	& o_{l,k} = \argmin_{i \in \{ 1, \ldots, L\}} \left( f_s(d^i_{l,\text{ref},k}) + s^i_{l,\text{lon},k} \right) \, ,\\
	& o_{r,k} = \argmin_{h \in \{ 1, \ldots, M\}} \left( f_s(d^h_{r,\text{ref},k}) + s^h_{r,\text{lon},k} \right) \, .
	\end{aligned}
\end{equation}\\
Consequently, the distances used in the MPC controller are
\begin{subequations}
	\begin{align*}
	& d_{l,\text{ref},k} = d^{o_{l,k}}_{l,\text{ref},k} \, ,\\
	& d_{r,\text{ref},k} = d^{o_{r,k}}_{l,\text{ref},k} \, ,\\
	& d_{l,\text{lon},k} = d^{o_{l,k}}_{l,\text{lon},k} \, ,\\
	& d_{r,\text{lon},k} = d^{o_{r,k}}_{r,\text{lon},k} \, .
	\end{align*}
\end{subequations}

The proposed approach is therefore able to handle as many objects as the perception/prediction layers can process.

\subsection{Diversifying the behavior for different classes of objects}

In case the perception system provides classified objects (e.g., car, bicycle, pedestrian), we may want to diversify the behavior depending on the object the AV deals with.
To this extent and since the number of classes is finite, extending the proposed framework is straightforward by adding as many constraints \eqref{constraint:f_s} as the number of classes and use different functions $f_s(d)$ (or the same function with different parameters) for different classes.

\section{RESULTS}
\label{sec:res}

We show experimental results obtained using the nuTonomy R\&D platform (Renault Zoe).
The experiments assume left-hand traffic (Singapore/UK regulations).


\subsection{Comparison with MPC without biasing}

\begin{figure}[thpb]
    \centering
  \subfloat[No biasing\label{subimg:no-biasing}]{%
       \includegraphics[width=0.45\linewidth]{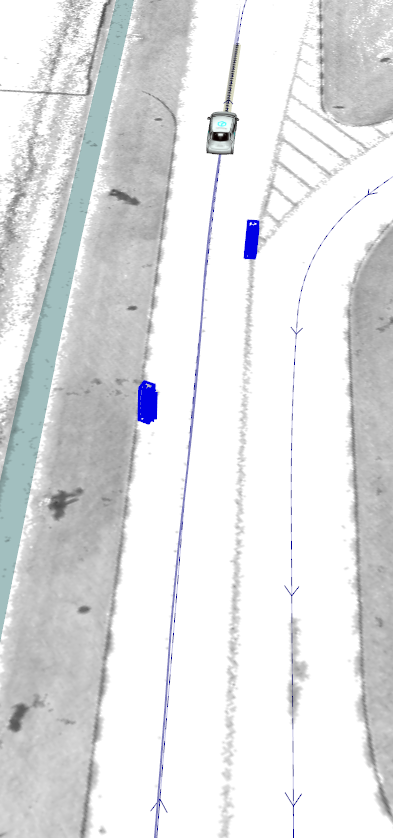}}
    \hfill
  \subfloat[Biasing\label{subimg:biasing}]{%
        \includegraphics[width=0.45\linewidth]{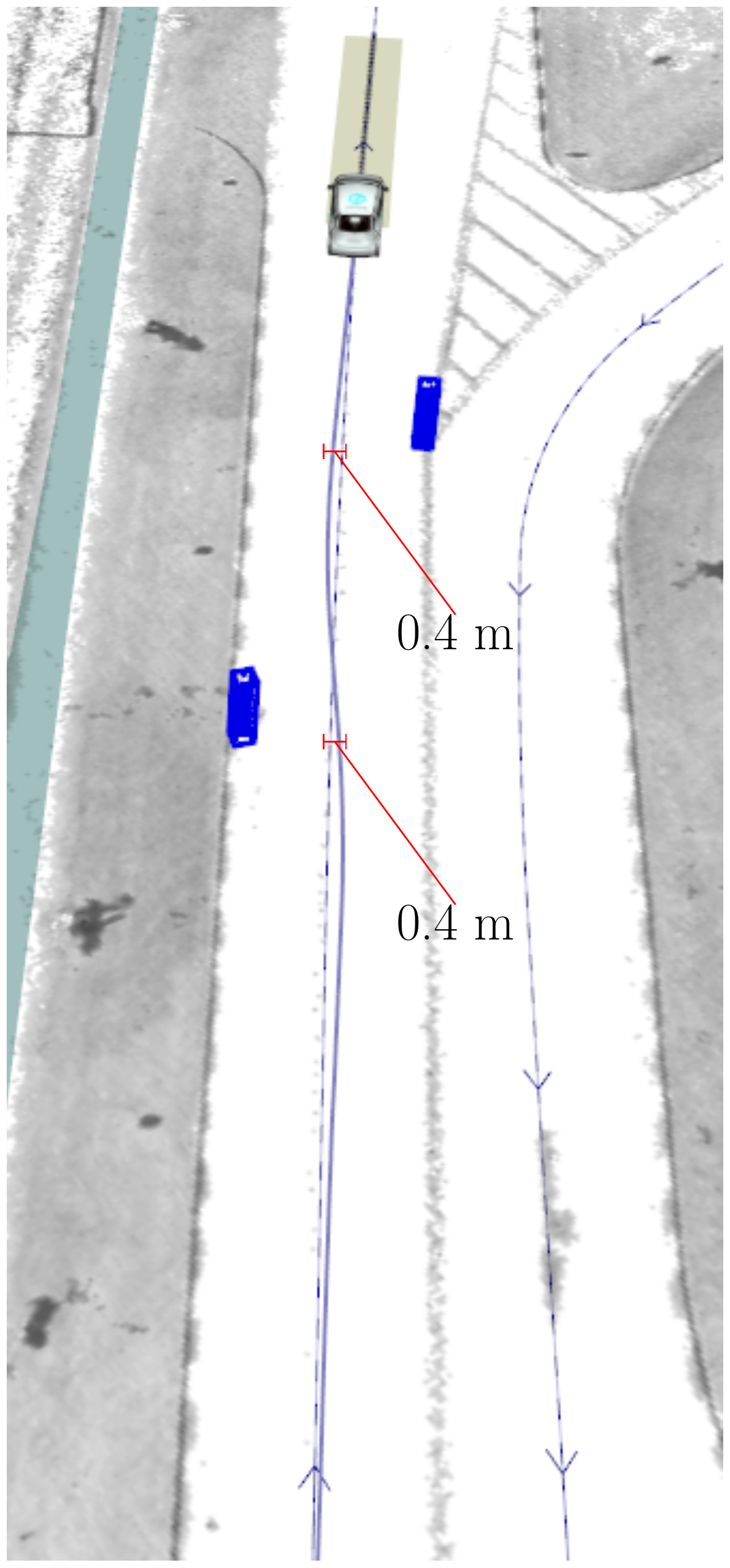}}
  \caption{Comparison of the normal MPC and the extended version for a scenario with two pedestrians (blue boxes). The continuous blue line represents the driven path.}
  \label{img:comparison}\vspace{-0.cm}
\end{figure}

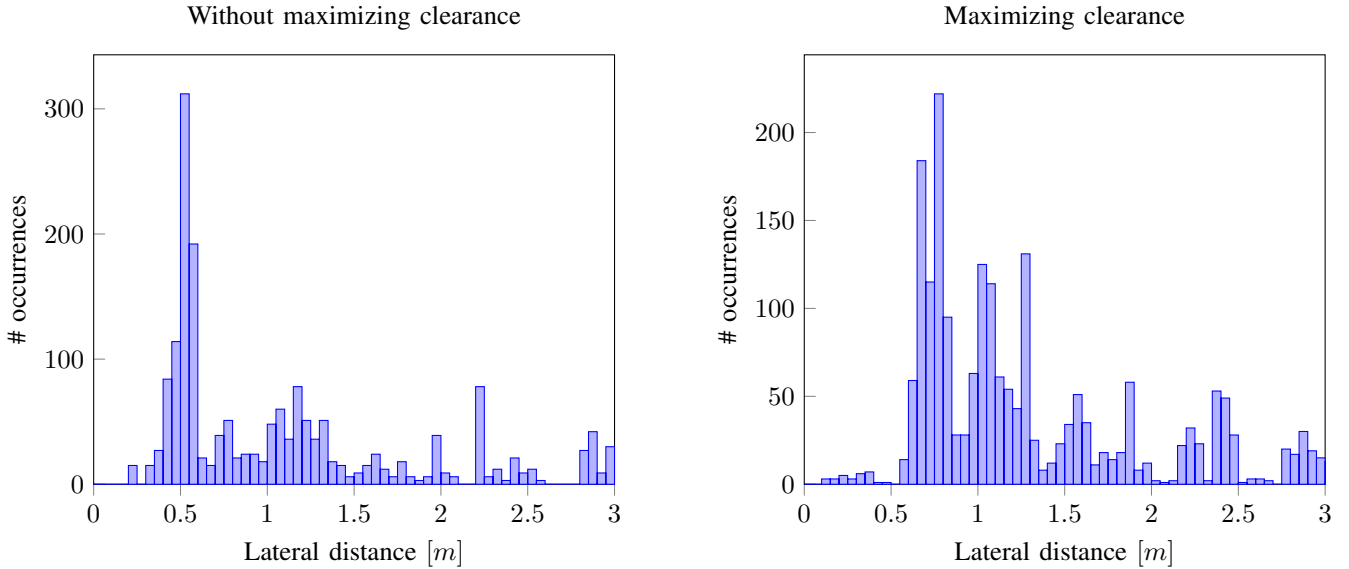
\begin{figure*}[thpb]
    \centering
  \subfloat{%
       \begin{tikzpicture}

\pgfmathsetmacro{\XendHist}{3}
\pgfmathsetmacro{\XendData}{4}
\pgfmathsetmacro{\Bins}{80}
\begin{axis}[
    ybar,
    xtick pos=left,
	ytick pos=left,
    ymin=0,
    xmin=0,
    xmax=\XendHist,
    xlabel={Lateral distance $[m]$},
    ylabel={\# occurrences},
    title={Without maximizing clearance}
]
\addplot +[
    hist={
        bins=\Bins,
        data min=0.0,
        data max=\XendData
    }   
] table [y index=0] {no_biasing.csv};

\end{axis}
\end{tikzpicture}}
    \hfill
  \subfloat{
        \begin{tikzpicture}

\pgfmathsetmacro{\XendHist}{3}
\pgfmathsetmacro{\XendData}{4}
\pgfmathsetmacro{\Bins}{80}
\begin{axis}[
    ybar,
    xtick pos=left,
	ytick pos=left,
    ymin=0,
    xmin=0,
    xmax=\XendHist,
    xlabel={Lateral distance $[m]$},
    ylabel={\# occurrences},
    title={Maximizing clearance}
]
\addplot +[
    hist={
        bins=\Bins,
        data min=0.0,
        data max=\XendData
    }
] table [y index=0] {biasing.csv};

\end{axis}
\end{tikzpicture}}
  \caption{Comparison of lateral clearance to other road agents during public road driving; the MPC with auxiliary goal (histogram on the right) consistently achieves higher lateral clearance.}
  \label{img:histograms}\vspace{-0.cm}
\end{figure*}

In the extended version, the parameters $a,b \geq 0$ in the clearance-dependent safety function \eqref{constraint:f_s} are tuned in simulation and verified through real car testing.
In case no road agent is present, the values $d_{\text{lat}}, d_{\text{lon}}$ are set to $\infty$ and $0$, respectively.
This causes the cost \eqref{mpccost_biasing} to equal $0$.
Similarly, if object $i$ results in  $f^i_s(d_{\text{ref}}) + s^i_{\text{lon}} > s_{\text{target}}$ (that is, is located very far), it is not accounted in Eq. \eqref{eq:dist_objects}.
The parameter $\alpha$ in Eq. \eqref{mpc_cost_full} was tuned by simulating different scenarios, and refined through real-world testing.

The prediction horizon employed is $N = 60$, corresponding to a time prediction of $\SI{3}{\second}$. MPC controllers with longer prediction horizons (i.e., $70-80$ time steps) were also tested, without particular benefits in terms of performance.

Fig. \ref{img:comparison} shows a real-world experiment consisting of our car driving with two pedestrians on the sides of the lane.
In the first picture \ref{subimg:no-biasing}, the simple MPC controller is used, and the car only follows the reference path. The closest lateral distance from its footprint to the pedestrian is about $\SI{1}{\meter}$.
In the second picture \ref{subimg:biasing}, the extended MPC controller is employed.
Additionally to following the reference trajectory, the car also maximizes lateral clearance to both pedestrians, and results in driving an S-shaped path (the blue line in the figure).
The closest lateral distance from its footprint to the pedestrian is about $\SI{1.4}{\meter}$, that is $40 \, \%$ larger than in the nominal case.
In terms of input magnitude, there are no substantial differences, resulting in the same comfort level for passengers (same yaw rate and acceleration).

A video showing experimental results can be found at https://vimeo.com/348076680/be686d1397.
For public road driving, the histograms in Fig. \ref{img:histograms} show the lateral distance given to other road agents (pedestrians, bicyclists, cars) for both the MPC controllers.
Such lateral distances are measured once the longitudinal displacement is equal to zero.
It is evident that the extended MPC shows an increased lateral clearance for a large number of occurrences.

\subsection{Numerical solvers used}

In order to numerically solve problem \eqref{MPC_formulation}, two off-the-shelf solvers tailored to MPC applications were tested, namely the ACADO \cite{Houska2011a} suite paired with QPOASES \cite{Ferreau2014}, and a proprietary Nonlinear Programming (NLP) solver.
Both frameworks provide the possibility to automatically discretize the continuous time system dynamics in Eq. \eqref{sys_dyn_ct}; the Explicit Runge-Kutta method of order 4 with a sampling time of $\SI{50}{\milli\second}$, is employed.
Even though both generate a high performing, tailored implementation of the MPC solver in C, they largely differ in terms of numerical algorithms employed.
The analysis and comparison of such numerical algorithms are beyond the scope of this paper.

\begin{table}[thpb]
  \label{tab:solvers}
  \caption{Solvetime of the two solvers tested, with and without biasing functionality. Horizon length $N = 60$.}
  \begin{center}
    \begin{tabular}{ | p{4cm} | P{1.3cm} | P{1.3cm} |}
      \hline
        & Average & Maximum \\
      \hline
        Proprietary NLP & $\SI{2.1}{\milli\second}$ & $\SI{3.7}{\milli\second}$ \\
      \hline
        Proprietary NLP (biasing) & $\SI{2.3}{\milli\second}$ & $\SI{5.4}{\milli\second}$ \\
      \hline
        ACADO+QPOASES & $\SI{1.9}{\milli\second}$ & $\SI{3.9}{\milli\second}$ \\
      \hline
        ACADO+QPOASES (biasing) & $\SI{2.0}{\milli\second}$ & $\SI{5.0}{\milli\second}$ \\
      \hline
    \end{tabular}
  \end{center}  
\end{table}
\vspace{-0.0cm}

\subsection{Solver comparison}

Table \ref{tab:solvers} shows the performance in terms of solvetime of the numerical solvers tested by comparing the traditional MPC tracking controller and its extended version with biasing functionality.
Both solvers were compiled with all the optimizations turned on and tested in identical conditions, running on a $4.6 \, \text{GHz i}7$ processor and sharing resources with other AV-relevant processes (perception, planning, etc.).

It is evident that both solvers are capable of finding a locally optimal solution to the problem within the time bounds provided (sampling time $T_s = \SI{50}{\milli\second}$), despite the extended MPC being a higher dimensional problem.

\section{CONCLUSION AND FUTURE WORK}

This paper presents an approach enabling the specification of additional tasks in an MPC tracking controller of an AV.
In particular, such task involves maximizing clearance to other road users, resulting in an increase of safety based on the TTR metric.
Moreover, extensive experimental results are showed as well as comparisons with traditional MPC tracking controllers.
The computational feasibility is assessed using two different numerical solvers.

Future development includes the implementation of the lane biasing functionality within a control framework commanding both speed and steering, as envisioned in \cite{nut-lane-biasing}.
The tractability of the resulting higher dimensional problem needs to be assessed, although the numerical results presented in Table \ref{tab:solvers} are encouraging.
Moreover, correlating the safety variable $s$ with the maximum attainable speed might be an interesting extension, allowing to drive faster in case of larger clearance and consequently mimicking observed human behaviors \cite{LLORCA2017302} and achieving a more human-like driving experience.

\section*{ACKNOWLEDGMENT}

The authors would like to thank the whole Aptiv team, for fruitful discussions and testing support.

\addtolength{\textheight}{-0cm}

\bibliographystyle{unsrt}
\bibliography{bib_ITSC}

\end{document}